\documentclass[11pt,a4paper]{article}
\usepackage[hyperref]{acl2021}
\usepackage{times}
\usepackage{latexsym}
\usepackage{tikz}
\usepackage{pgfplots}
\pgfplotsset{compat=1.7}
\usepackage{comment}
\usepackage{multirow}

\usepackage{microtype}

\aclfinalcopy 

\title{Explainable Tsetlin Machine framework for fake news detection with credibility score assessment}

\author{Bimal Bhattarai \\
  University of Agder \\
  \texttt{bimal.bhattarai@uia.no} \\\And
  Ole-Christoffer Granmo \\
  University of Agder \\
  \texttt{ole.granmo@uia.no} \\ \And
  Lei Jiao \\
  University of Agder \\
  \texttt{lei.jiao@uia.no} \\}

\date{}

\begin{document}
\maketitle
\begin{abstract}
The proliferation of fake news, i.e., news intentionally spread for misinformation, poses a threat to individuals and society. Despite various fact-checking websites such as PolitiFact, robust detection techniques are required to deal with the increase in fake news. Several deep learning models show promising results for fake news classification, however, their black-box nature makes it difficult to explain their classification decisions and quality-assure the models. We here address this problem by proposing a novel interpretable fake news detection framework based on the recently introduced Tsetlin Machine (TM). In brief, we utilize the conjunctive clauses of the TM to capture lexical and semantic properties of both true and fake news text. Further, we use the clause ensembles to calculate the credibility of fake news. For evaluation, we conduct experiments on two publicly available datasets, PolitiFact and GossipCop, and demonstrate that the TM framework significantly outperforms previously published baselines by at least $5\%$ in terms of accuracy, with the added benefit of an interpretable logic-based representation. Further, our approach provides higher F1-score than BERT and XLNet, however, we obtain slightly lower accuracy. We finally present a case study on our model's explainability, demonstrating how it decomposes into meaningful words and their negations.
\end{abstract}
\section{Introduction}
Social media platforms on the Internet have become an integral part of everyday life, and more and more people obtain news from social rather than traditional media, such as newspapers. Key drivers include cost-effectiveness, freedom to comment, and shareability among friends. Despite these advantages, social media exposes people to abundant misinformation. Fake news stories are particularly problematic as they
seek to deceive people for political and financial gain \cite{32}.

In recent years, we have witnessed extensive growth of fake news in social media, spread across news blogs, Twitter, and other social platforms. At present, most online misinformation is manually written \cite{2}. However, natural language models like GPT-3 enable automatic generation of realistic-looking fake news, which may accelerate future growth. Such growth is problematic as most people nowadays digest news stories from social media and news blogs \cite{1}. Indeed, the spread of fake news poses a severe threat to journalism, individuals, and society. It has the potential of breaking societal belief systems and encourages biases and false hopes. For instance, fake news related to religion or gender inequality can produce harmful prejudices. Fake news can also trigger violence or conflict. When people are frequently exposed to fake news, they tend to distrust real news, affecting their ability to distinguish between truth and untruth. To reduce these negative impacts, it is critical to develop methods that can automatically expose fake news.

Fake news detection introduces various challenging research problems. By design, fake news intentionally deceives the recipient. It is therefore difficult to detect fake news based on linguistic content, style, and diverseness. For example, fake news may narrate actual events and context to support false claims \cite{3}. Thus, other than hand-crafted and data-specific features, we need to employ a knowledge base of linguistic patterns for effective detection.  Training fake news classifiers on crowdsourced data may further provide a poor fit for future news events. Fake news is emerging continuously, quickly rendering previous textual content obsolete. Accordingly, some studies, such as \cite{4}, have tried to incorporate social context and hierarchical neural networks using attention to uncover more lasting semantic patterns.

Despite significant advances in deep learning-based techniques for fake news detection, few approaches can explain their classification decisions. Currently, knowledge on the dynamics underlying fake news is lacking. Thus, explaining why certain news items are considered fake may uncover new understanding. Making the reasons for decisions transparent also facilitates discovering and rectifying model weaknesses. To the best of our knowledge, previous research has not yet addressed explainability in fake news detection.

\textbf{Paper contributions:} In this paper, we propose an explainable framework for fake news detection built using the Tsetlin Machine (TM).  Our TM-based framework captures the frequent patterns that characterize real news, distilling both linguistic and semantic patterns unique for fake news stories. The resulting model constitutes a global interpretation of fake news. To provide a more refined view on individual fake news (local interpretability), we also propose a credibility score for measuring the credibility of news.  Finally, our framework allows the practitioner to see what features are critical for making news fake (global interpretability).

\textbf{Paper organization:} The remainder of the paper is organized as follows. In Section \ref{RW}, we briefly summarize related work. We then detail our explainable framework in Section \ref{EF}. In Section \ref{EE}, the datasets and experimental configurations are presented. We analyze our empirical results in Section \ref{RD}, before we conclude the work in the last section.  

\section{Related Work}\label{RW}

The problem of detecting deception is not new to natural language processing \cite{2}. Significant  application domains include detecting false online advertising, fake consumer reviews, and spam emails \cite {24} \cite{25}. The detection of \emph{fake news} focuses on uncovering spread of misleading news articles \cite {15} \cite{16}. Typical detection techniques use either text-based linguistic features \cite{17} or visual features \cite{18}. Overall, fake news detection methods fall into two groups: knowledge-based models based on fact-checking news articles using external sources \cite {21}, and style-based models, which leverage linguistic features capturing writing style \cite{22}. Many studies such as \cite{19} \cite{20} \cite{14} incorporate publicly available datasets, providing a basis for detailed analysis of fake news and detection methods. 

Recently, deep learning-based latent representation of text has significantly improved fake news classification accuracy \cite{23}. However, the latent representations are generally difficult to interpret, providing limited insight into the nature of fake news. In \cite {26}, the authors introduced features based on social context, obtained from the profiles of users and their activity patterns.  Other approaches depend upon social platform-specific features such as likes, tweets, and retweets for supervised learning \cite{27} \cite{28}.

\begin{figure*}
    \centering
    \includegraphics[width=\textwidth]{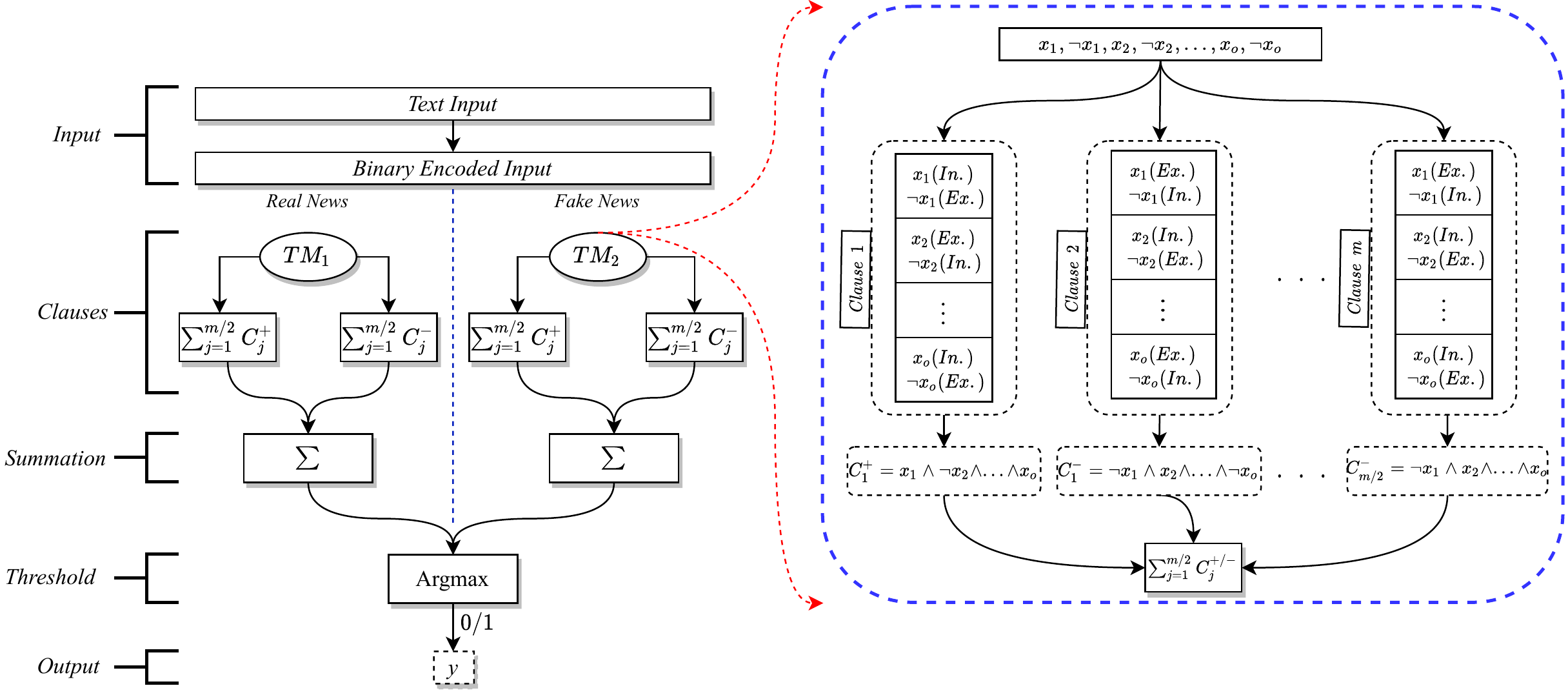}
    \caption{ Interpretable Tsetlin machine architecture.}
    \label{fig:tsetlin_architecture}
\end{figure*}

While the progress on detecting fake news has been significant, limited effort has been directed towards interpretability.   Existing deep learning methods generally extract features to train classifiers without giving any interpretable explanation. This lack of transparency makes them black boxes when it comes to understandability \cite {29}.
In this paper, we propose a novel fake news detection method that builds upon the TM \cite{5}. The TM is a recent approach to pattern classification, regression, and novelty detection \cite{abeyrathna2021parallel,rohan2021AAAI,abeyrathna2020nonlinear, 33} that attempts to bridge the present gap between interpretability and accuracy in state-of-the-art machine learning. By using the AND-rules of the TM to capture lexical and semantic properties of fake news, our aim is to improve the performance of fake news detection. More importantly, our framework is intrinsically interpretable, both globally, at the model level, and locally for the individual false news predictions.

\section{Explainable Fake News Detection Framework}\label{EF}
In this section, we present the details of our TM-based fake news detection framework. 

\subsection{TM Architecture}\label{architecture}
The TM consists of a team of two-action Tsetlin Automata (TAs) with $2N$ states. Each TA performs either action ``Include" (in State $1$ to $N$) or action ``Exclude" (in State $N$ to $2N$). Jointly, the actions specify a pattern recognition task. Action ``Include" incorporates a specific sub-pattern, while action ``Exclude" rejects the sub-pattern. The TAs are updated based on iterative feedback in the form of rewards or penalties. The feedback signals how well the pattern recognition task is solved, with the intent of maximizing classification accuracy. Rewards reinforce the actions performed by the TAs, while penalties suppress the actions.

By orchestrating the TA team with rewards and penalties, a TM can capture frequent and discriminative patterns from labeled training data. Each pattern is represented as a conjunctive clause in propositional logic, based on the human-interpretable disjunctive normal form \cite{valiant1984learnable}. That is, the TM produces AND-rules formed as conjunctions of propositional variables and their negations.

\begin{figure*}
    \centering
    \includegraphics[width=\textwidth]{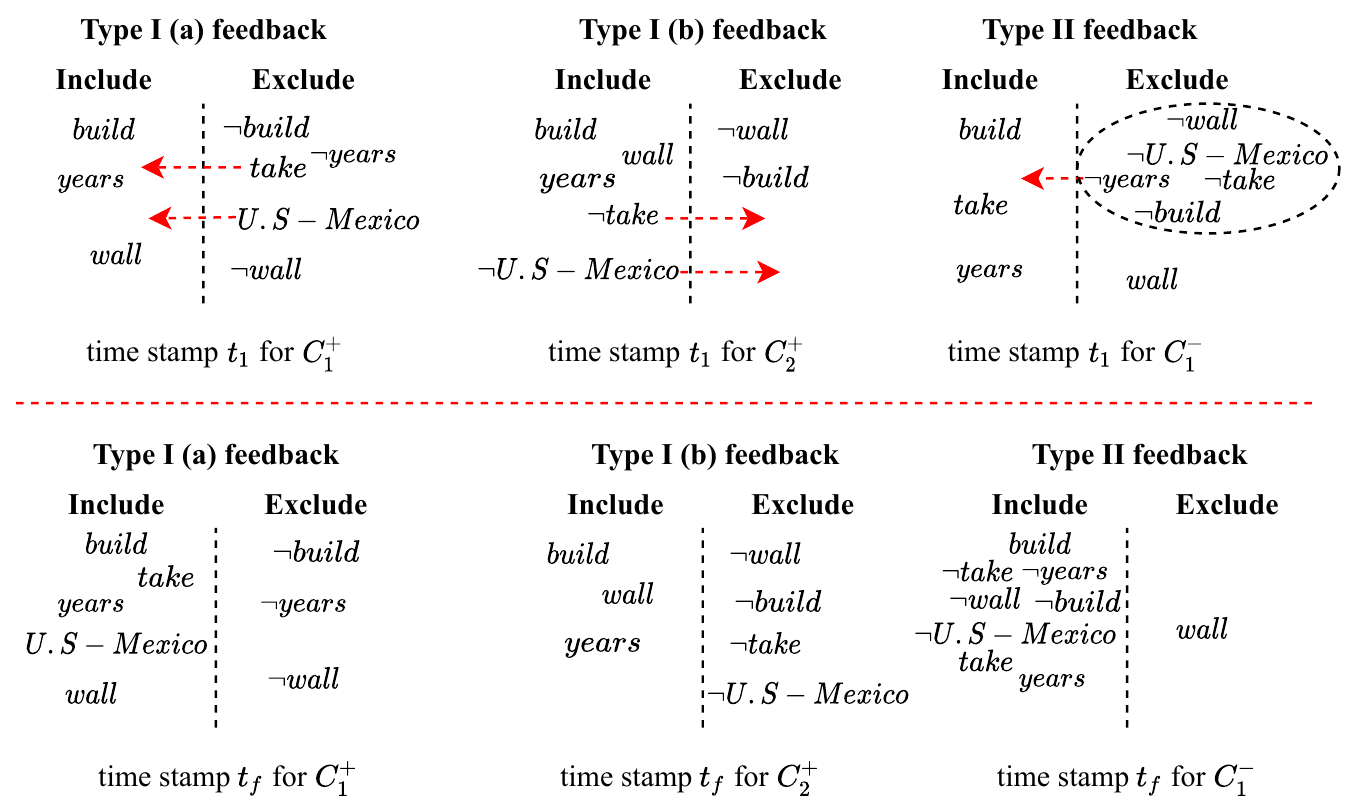}
    \caption{Visualization of Tsetlin machine learning.}
    \label{fig:tsetlin_learning}
\end{figure*}

A two-class TM structure is shown in Figure~\ref{fig:tsetlin_architecture}, consisting of two separate TMs ($TM_1$ and $TM_2$). As seen in the Input-step, each TM takes a Boolean (propositional) vector $X= (x_1, \ldots, x_o)$, $x_k\in\{0,1\}$, $k\in\{1,\ldots, o\}$ as input, which is obtained by booleanizing the text input as suggested in \cite{6, rohan2021AAAI}. That is, the text input is modelled as a set of words, with each Boolean input signaling the presence or absence of a specific word. From the input vector, we obtain $2o$ literals $L = (l_1, l_2, \ldots, l_{2o})$. The literals consist of the inputs $x_k$ and their negated counterparts $\bar{x}_k = \lnot x_k = 1-x_k$, i.e.,  $L = (x_1, \ldots, x_n, \neg x_1, \ldots, \neg x_n)$.

A TM forms patterns using $m$ conjunctive clauses $C_j$ (Figure~\ref{fig:tsetlin_architecture} -- Clauses). How the patterns relate to the two output classes ($y=0$ and $y=1$) is captured by assigning polarities to the clauses. Positive polarity is assigned to one half of the clauses, denoted by $C_j^+$. These are to capture patterns for the target class ($y=0$ for $TM_1$ and $y=1$ for $TM_2$). Negative polarity is assigned to the other half, denoted by $C_j^-$. Negative polarity clauses are to capture patterns for the non-target class ($y=1$ for $TM_1$ and $y=0$ for $TM_2$). In effect, the positive polarity clauses vote for the input belonging to the target class, while negative polarity clauses vote against the target class.

Any clause $C_j^+$ for a certain target class is formed by ANDing a subset $L_j^+ \subseteq L$ of the literal set, written as:
\begin{equation}
	C_j^+ (X)=\bigwedge_{l_k \in L_j^+} l_k = \prod_{l_k \in L_j^+} l_k,
\end{equation} 
where $j = (1, \ldots, m/2)$ denotes the clause index, and the superscript decides the polarity of the clause.
For instance, the clause $C_j^+(X) = x_1 x_2$ consists of the literals $L_j^+ = \{x_1, x_2\}$, and it outputs $1$ if both of the literals are $1$-valued. Similarly, we have clauses $C^-_j$ for the non-target class.

The final classification decision is done after summing up the clause outputs (Summation-step in the figure). That is, the negative outputs are substracted from the positive outputs. Employing a single TM, the sum is then thresholded using the unit step function $u$, $u(v) = 1 ~\mathbf{if}~ v \ge 0 ~\mathbf{else}~ 0$, as shown in Eq.~(\ref{total_output}):
\begin{equation}
    \hat{y} = u\left(\sum_{j=1}^{m/2} C_j^+(X) - \sum_{j=1}^{m/2} C_j^-(X)\right).
     \label{total_output}
\end{equation}
The summation of clause outputs produces an adaptive ensemble effect designed to  help dealing with noisy and diverse data \cite {5}. For example, the classifier $\hat{y} = u\left(x_1 \bar{x}_2 + \bar{x}_1 x_2 - x_1 x_2 - \bar{x}_1 \bar{x}_2\right)$ captures the XOR-relation.

With multi-class problems, the classification is instead performed using the $\mathrm{argmax}$ operator, as shown in the figure. Then the target class of the TM with the largest vote sum is given as output. 

\subsection{Interpretable Learning Process}
As introduced briefly above, the conjunctive clauses in a TM are formed by a collection of TAs. Each TA decides whether to ``Include" or ``Exclude" a certain literal in a specific clause based on reinforcement, i.e., rewards, penalties, or inaction feedback. The reinforcement depends on six factors: (1) target output ($y = 0$ or $y = 1$), (2) clause polarity, (3) clause output ($C_j = 0$ or $1$), (4) literals value (i.e., $x =1$, and $\neg x = 0$), (5) vote sum, and (6) the current state of the TA.

The TM learning process carefully guides the TAs to converge towards optimal decisions. To this end, the TM organizes the feedback it gives to the TAs into two feedback types. Type I feedback is designed to produce frequent patterns, combat false negatives, and make clauses evaluate to $1$. Type I feedback is given to positive polarity clauses when $y = 1$ and to negative polarity clauses when $y=0$. Type II feedback, on the other hand,  increases the discriminating power of the patterns, suppresses false positives, and makes clauses evaluate to $0$. Type II feedback is given to positive polarity clauses when $y=0$ and to negative polarity clauses when $y=1$. 

The feedback is further regulated by the sum of votes $v$ for each output class. That is, the voting sum is compared against a voting margin  $T$, which is employed to guide distinct clauses to learn different sub-patterns. The details of the learning process can be found in \cite{5}. \par

We use the following sentence as an example to show how the inference and learning process can be interpreted: $X$= [``Building a wall on the U.S-Mexico border will take literally years.''], with output target ``true news'' i.e., $y=1$. First, the input is tokenized and negated: $X = [``build"=1, ``\neg build"=0, ``wall"=1,``\neg wall"=0, ``U.S-Mexico"=1, $ $``\neg U.S -Mexico"=0, $ $``take"=1, $ $``\neg take"=0, $ $ ``years"=1,$ $``\neg years"=0]$. We consider two positive polarity clauses and one negative polarity one (i.e., $C_1^+$, $C_2^+$, and  $C_1^-$) to show different feedback conditions occurring while learning the propositional rules.

The learning dynamics are illustrated in Figure~\ref{fig:tsetlin_learning}. The upper part of the figure shows the current configuration of clauses and TA states for three different scenarios. On the left side of the vertical bar, the literals are included in the clause, and on the right side of the bar, the literals are excluded. The upper part also shows how the three different kinds of feedback modifies the states of the TAs, either reinforcing ``Include" or ``Exclude". The lower part depicts the new clause configurations, after the reinforcement has been persistently applied.

For the first time stamp in the figure, we assume the clauses are initialized randomly as follows: $C_1^+ = (build \wedge wall \wedge  years)$,  $C_2^+= (build \wedge wall \wedge years \wedge \neg take \wedge \neg U.S-Mexico)$, and $C_1^-= ( build \wedge  take \wedge years)$. Since clause $C_1^+$ consists of non-negated literals of value $1$ in the input $X$, the output becomes $1$ (because  $C_1^+ =  1 \wedge 1 \wedge 1)$. This invokes the condition ($C_1^+ =1$ and $y=1$). So, as shown in upper left of Figure~\ref{fig:tsetlin_learning}, when the actual output is $1$ and the clause output is $1$, Type~I~(a) feedback is used to reinforce the $Include$ action. Therefore, literals such as $U.S-Mexico$ and $take$ are eventually included, because they appear in $X$. This process makes the clause $C_1^+$ eventually have one sub-pattern captured for $y=1$, as shown on the lower left side of Figure~\ref{fig:tsetlin_learning}. I.e., $take$ and $U.S-Mexico$ have been included in the clause. \par

Similarly, in $C_2^+$, the clause output is $0$ because of the negated inputs $\neg take$ and $\neg U.S-Mexico$ (i.e., $C_2^+= 1 \wedge 1\wedge 1 \wedge 0 \wedge 0$). This invokes the condition ($C_2^+=0$ and $y=1$). Here, so-called Type~I~(b) feedback is used to reinforce $Exclude$ actions. In this example, literals such as $\neg take$ and $\neg U.S-Mexico$ are eventually excluded from $C_2^+$ as shown in the middle part of Figure  \ref{fig:tsetlin_learning}, thus establishing another sub-pattern for $C_2^+$.

Type~II feedback (right side of figure) only occurs when $y=0$ (for positive polarity clauses) and $y=1$ (for negative polarity clauses). We here use the negative polarity clause $C_1^-$ to demonstrate the effect of Type~II feedback. This type of feedback is triggered when the clause output is $1$. The goal is now to make the output of the affected clause change from $1$ to $0$ by adding $0$-valued inputs. Type~II feedback can be observed on the right side of Figure \ref{fig:tsetlin_learning}, where all negated literals finally are included in $C_1^-$ to ensure that the clause outputs~$0$. 

\subsection{Problem Statement for Fake News Detection}

We now proceed with defining the fake news detection problem formally. Let $\mathcal{A}$ be a news article with $s_i$ sentences, where $i= 1$ to $\mathcal{S}$ and $\mathcal{S}$ is the total number of sentences. Each sentence can be written as $s_i=(w_1^i, w_2^i, \ldots, w_\mathcal{W}^i)$, consisting of $\mathcal{W}$ words. Given the booleanized input vector $X \in \{0,1\}^o$, the fake news classification is a Boolean classification problem, where we predict whether the news article $\mathcal{A}$ is fake or not, i.e., $\mathcal{F} : X \rightarrow y \in \{0, 1\}$.
%
We also aim to learn the credibility of news by formulating a TM-based credibility score that measures how check-worthy the news is. With this addition, the classifier function can be written as $\mathcal{F} : X \rightarrow (y, \mathcal{Q})$, where $y$ is a classification output and $\mathcal{Q}$ is the credibility score.

\subsection{Credibility Assessment} \label{credibility_assessment}

The credibility score can be calculated as follows. Firstly, the TM architecture in Section \ref{architecture} is slightly tweaked for the score generation. In the architecture, instead of identifying the class with the largest voting sum using \textit{Argmax}, we obtain the raw score from both the output classes. The raw score is generated using clauses, which represent frequent lexical and semantic patterns that characterizes the respective classes. Therefore, the score contains information on how the input resembles the patterns captured by the clauses. We thus use this score to measure the credibility of news. For instance, consider the  vote ratios of 2:1 and 10:1 for two classes (i.e., real vs. fake news) obtained from two different inputs. Then the first class wins the majority vote in both cases, however,  the 10:1 vote ratio suggests higher credibility than for the input that gave a ratio of  2:1.\par

To normalize the credibility score so that it falls between $0$ and $1$, we apply the logistic function to the formula in Eq.~(\ref{total_output}) as shown in Eq.~(\ref{logistic_function}).
\begin{equation}
    \mathcal{Q}_i = \frac{1}{1 + \exp ^ {-k \left( v_i^F - v_i^T \right) }}.
    \label{logistic_function}
\end{equation}
Above, $\mathcal{Q}_i$ is the credibility score of the $i^{th}$ sample, and $k$ is the logistic growth rate.  $v_i^F $ and $v_i^T$ are the total sum of votes collected by clauses for fake and true news respectively. Here, $k$ is a user configurable parameter deciding the slope of the function. For example, consider the scores (43, -47) and (124, -177), both predicting fake class.  The credibility scores obtained from Eq.~(\ref{logistic_function}) with $k=0.012$ are 0.74 and 0.97, which shows that the second news is more credible than the first one. 

\section{Experiment Setup}\label{EE}
In this section, we present the experiment configurations we use to evaluate our proposed TM framework.

\subsection{Datasets}
For evaluation, we adopt the publicly available fake news detection data repository FakeNewsNet \cite{7}. The repository consists of news content from different fact-checking websites, including social context and dynamic information. We here use news content annotated with labels by professional journalists from the fact-checking websites \textit{PolitiFact} and \textit{GossipCop}.  PolitiFact is a fact-checking website that focuses on U.S political news. We extract news articles published till 2017. GossipCop focuses on fact-checking of entertainment news collected from various medias. While GossipCop has more fake news articles than PolitiFact, PolitiFact is more balanced, as shown in Table {\ref {table1}}.

\subsection{Preprocessing}
The relevant news content and the associated labels, i.e., True or False news, are extracted from both of the datasets. The preprocessing steps include cleaning, tokenization, lemmatization, and feature selection. The cleaning is done by removing less important information such as hyperlinks, stop words, punctuation, and emojis. The datasets are then booleanized into a sparse matrix for the TM to process. The matrix is obtained by building a vocabulary consisting of all unique  words in the dataset, followed by encoding the input as a set of words using that vocabulary. To reduce the sparseness of the input, we adopt two methods: 1) Chi-squared test statistics as a feature selection technique, and 2) selecting the most frequent literals from the dataset. The experiment is performed using both methods, and the best results are included. For the PolitiFact and GossipCop datasets, we selected the $20~000$ and $25~000$ most significant features, respectively.

\begin{table}
    \begin{center}
    \begin{tabular}{ l|c|c|c } \hline
    Dataset & \#Real &\#Fake &\#Total \\ \hline
    \textit{PolitiFact} & 563 &391 & 954\\ \hline
    \textit{GossipCop} &15,338 & 4,895 & 20,233\\ \hline
    \end{tabular}
    \caption{Dataset statistics. }
    \label{table1}
    \end{center}
    \vspace{-0.5cm}
\end{table}

\subsection{Baseline}
We first summarize the state-of-the-art of the fake news detection approaches. 

\begin{itemize}
	\item RST \cite{8}:  Rhetorical Structure Theory (RST) represents the relationship between words in a document by building a tree structure. It extracts the news style features from a bag-of-words by mapping them into a latent feature representation. 
	\item LIWC \cite{9}: Linguistic Inquiry and Word Count (LIWC) is used to extract and learn features from psycholinguistic and deception categories. 
	\item HAN \cite{10}: HAN uses a hierarchical attention neural network (HAN) for embedding word-level attention on each sentence and sentence-level attention on news content, for fake news detection.
	\item CNN-text \cite{11}: CNN-text utilizes convolutional neural network (CNN) with pre-trained word vectors for sentence-level classification. The model can capture different granularities of text features from news articles via multiple convolution filters. 
	\item LSTM-ATT \cite{12}: LSTM-ATT utilizes long short term memory (LSTM) with an attention mechanism. The model takes a 300-dimensional vector representation of news articles as input to a two-layer LSTM for fake news detection.
	\item RoBERTa-MWSS: The Multiple Sources of Weak Social Supervision (MWSS) approach, built upon RoBERTa \cite{35}, was proposed in \cite{34}.
	\item BERT \cite{36}: Bidirectional Encoder Representations from Transformers (BERT) is a Transformer-based model which contains an encoder with 12 transformer blocks, self-attention heads and hidden shape size of 768.
	\item XLNet \cite{37}: XLNet is a generalized autoregressive pretraining model that integrates autoencoding and a segment recurrence mechanism from transformers.
\end{itemize}

In addition, we compare our results with other machine learning baseline models such as logistic regression (LR), naive Bayes, support vector machines (SVM), and random forest (RF). For a fair comparison, we select the methods that only extract textual features from news articles. The performance for these baseline models has been reported in \cite {13} and \cite{14}.

\subsection{Training and Testing}
We use a random training-test split of  75\%/25\%. The classification results is obtained on the test set. The process is repeated five times, and the average accuracy and F1 scores are reported. To provide robust results, we calculated an ensemble average by first take the average of $50$ stable epochs, followed by taking the average of the resulting five averages. We run our TM for $200$ epochs with hyperparameter configuration of $10~000$ clauses, a threshold $T$ of $200$, and sensitivity $s$ of $25.0$. To ensure a fair comparison of the results obtained by other machine learning algorithms, we use the Python Scikit-learn framework that includes LR, SVM and Naïve Bayes, using default parameter settings.

\section{Results and Discussion}\label{RD}
\label{result_discussion}
We now compare our framework with above mentioned baseline models.

\subsection{Comparison with State-of-the-Art}

The experimental results are shown in Table \ref{results}. Clearly, our model outperforms several of the other baseline models in terms of accuracy and F1 score.  
\begin{table}[ht!]
\centering

\small
    \begin{tabular}{l|cc|cc}
    \hline
    \multirow{2}{*}{Models} & \multicolumn{2}{c}{ PolitiFact } & \multicolumn{2}{c}{ GossipCop }\\ \cline{2-5}
    & Acc. & F1 & Acc. & F1\\
    \hline
    RST     & 0.607  &   0.569    &    0.531   &   0.512  \\
    LIWC      & 0.769   &  0.818    &   0.736    &  0.572 \\
    HAN       & 0.837  &   0.860    &   0.742    & 0.672  \\
    CNN-text  & 0.653   &   0.760    &   0.739    & 0.569 \\
    LSTM-ATT  & 0.833   &  0.836    &   0.793    &  0.798 \\
    LR  & 0.642   &   0.633    & 0.648  & 0.646  \\
    SVM  & 0.580   &   0.659    & 0.497  & 0.595   \\
    Naïve Bayes & 0.617   &   0.651    & 0.624  & 0.649   \\
    RoBERTa-MWSS & 0.825   &   0.805    & 0.803  & 0.807   \\
    BERT & 0.88 & 0.87 & 0.85 & 0.79 \\
    XLNet & \textbf{0.895} & 0.90 & \textbf{0.855} & 0.78 \\
    \textbf{TM} & 0.871 & \textbf{0.901} & 0.842 & \textbf{0.896} \\
    \hline
    \end{tabular}
    \caption{Performance comparison of our model with 8 baseline models.}
 \label{results}    
\end{table}

One can further observe that HAN outperforms LIWC, CNN-text, and RST for both datasets. This is arguably because HAN can capture the syntactic and semantic rules using hierarchical attention to detect fake news. Similarly, the LIWC performs better than RST. One possible explanation for this is that LIWC can capture the linguistic features in news articles based on words that denote psycholinguistic characteristics. The LSTM-ATT, which has extensive preprocessing using count features and sentiment features along with hyperparameter tuning \cite {12}, has similar performance compared with HAN in PolitiFact, however, outperforms HAN on GossipCop. One reason for this can be that the attention mechanism is able to capture the relevant representation of the input. In addition, we see that machine learning models such as LR, SVM, and Naive Bayes are not very effective in neither of the datasets.\par

The recent transformer-based models BERT and XLNet outperform our model in terms of accuracy. Our TM-based approach obtains slightly lower accuracy, achieving $87.1\%$ for PolitiFact and $84.2 \%$ for GossipCop. However, the F1 score for politiFact is $0.90$, whereas for GossipCop it is $0.89$, which is marginally better than XLNet and BERT. Thus, we achieve state-of-the-art performance with respect to F1-score overall, and with respect to accuracy when comparing with interpretable approaches. Since GossipCop is unbalanced with sample ratio of 3:1 for real and fake news, we submit that the F1 score is a more appropriate performance measure than accuracy. Overall, our model is much simpler than the deep learning models because we do not use any pre-trained embeddings for preprocessing. This also helps in making our model more transparent, interpretable, and explainable.\par

\begin{figure}
    \centering
    \includegraphics[width=7cm]{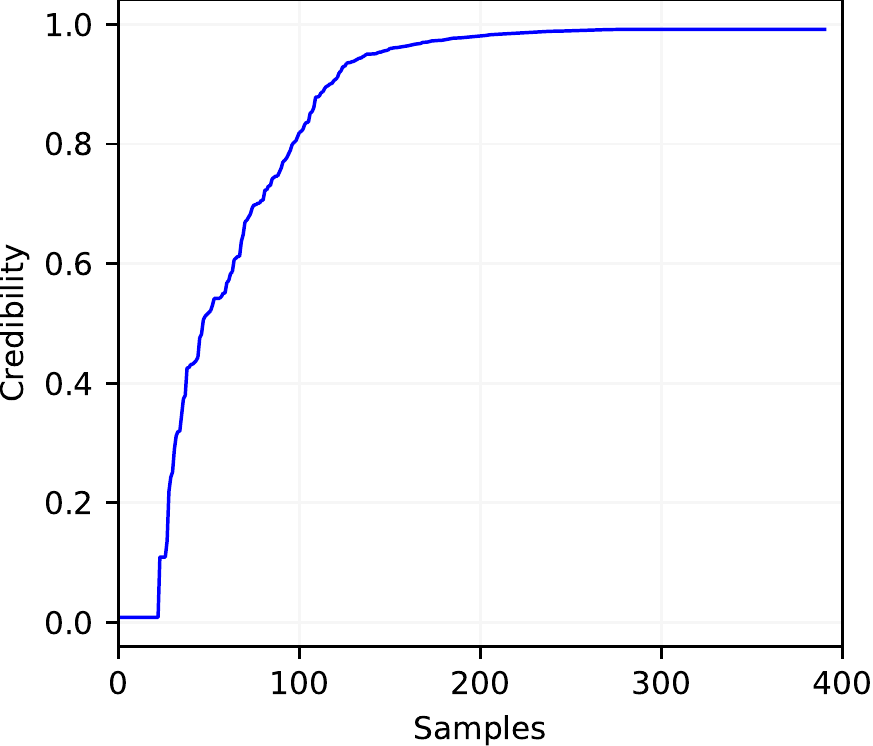}
    \caption{ Credibility assessment for fake news in PolitiFact.}
    \label{fig:politi_credibility}
\end{figure}

\begin{figure}
    \centering
    \includegraphics[width=7cm]{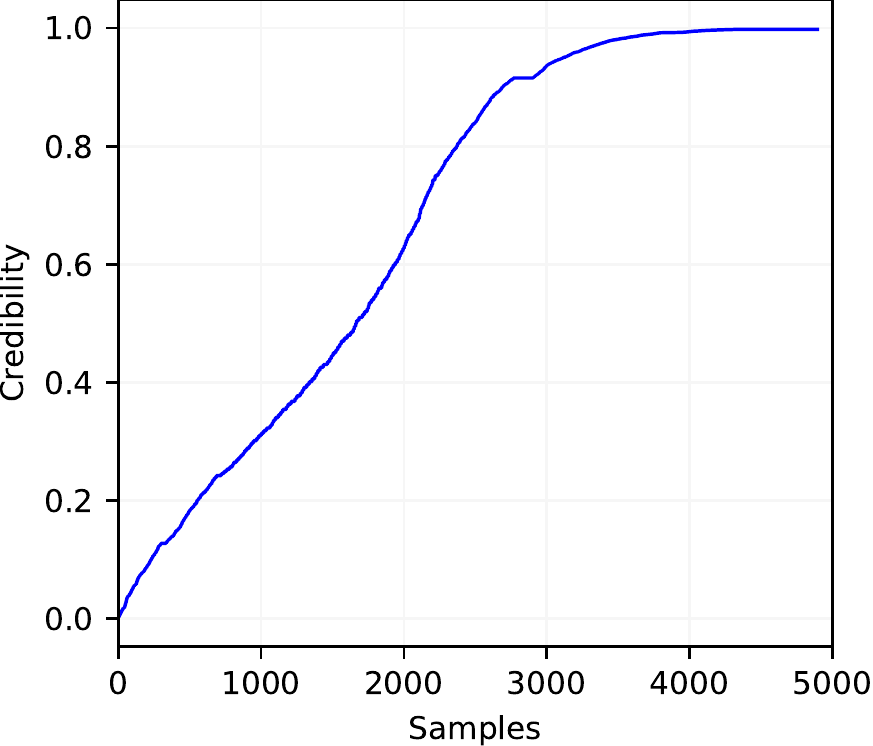}
    \caption{ Credibility assessment for fake news in GossipCop.}
    \label{fig:gossipcop_credibility}
    \vspace{-0.5cm}
\end{figure}

The credibility assessment is performed as described in Subsection \ref{credibility_assessment}. The Figs.~\ref{fig:politi_credibility} and \ref{fig:gossipcop_credibility} show the credibility scores for a fake news sample from PolitiFact and GossipCop, resepectively. As seen, the fake news can be ranked quite distinctly. This facilitates manual checking according to credibility, allowing users to focus on the fake news articles with highest scores. However, when we observe the soft scores e.g. obtained from XLNet, most of the samples are given rather extreme scores. That is, classifications are in general submitted with very high confidence. This is in contrast to the more cautious and diverse credibility scores produced by TM clause voting. If we for instance set a credibility score threshold of $0.8$ in TM, we narrow the selection of fake news down to around $300$ out of $400$ for PolitiFact and to around $2~800$ out of $4~895$ for GossipCop.

\begin{table*}[ht!]
\centering

\small
    \begin{tabular}{llll|llll}
    \hline
    \multicolumn{8}{c}{ PolitiFact }\\ \cline{1-8}
    \multicolumn{4}{c}{Fake} & \multicolumn{4}{c}{True}\\ \cline{1-8}
     Plain & times & Negated & times & Plain & times & Negated & times \\
    \hline
     trump &   297    &    candidate   &   529 & congress  &   136    &    trump   &   1252 \\
     said   &  290    &   debate    &  413 & tax & 104 & profession & 1226\\
     comment  &   112    &   civil   & 410 & support & 70 & navigate & 1223\\
     donald   &   110    &   reform    & 369 & senate & 64& hackings & 1218\\
     story   &  78    &   congress    &  365 & president & 60& reported & 1216\\
     medium   &   63    & iraq  & 361  & economic & 57 & arrest  & 1222\\
     president   &   48    & lawsuit  & 351 & americans & 49  & camps & 1206\\
    reported   &   45   & secretary  & 348 & candidate & 48 &  investigation & 1159\\
    investigation   &   38   & tax  & 332 & debate & 44 &  medium & 1152\\
    domain   &   34   & economy  & 321 & federal & 41  & domain & 1153\\
    \hline
    \end{tabular}
    \caption{Top ten Literals captured by clauses of TM for PolitiFact.}
 \label{explainability_table_politifact} 
\end{table*}

\begin{table*}[ht!]
\centering

\small
    \begin{tabular}{llll|llll}
    \hline
    \multicolumn{8}{c}{ GossipCop }\\ \cline{1-8}
    \multicolumn{4}{c}{Fake} & \multicolumn{4}{c}{True}\\ \cline{1-8}
     Plain & times & Negated & times & Plain & times & Negated & times \\
    \hline
     source &   357    &    stream   &   794 & season  &   150    &    insider   &   918 \\
     insider   &  152   &   aggregate    &  767 & show & 103 & source & 802\\
     rumors  &   86   &   bold    & 723  & series & 79& hollywood & 802\\
     hollywood   &   80    &   refreshing   & 722 & like & 78& radar & 646\\
     gossip   &  49    &   castmates    &  721 & feature & 70& cop & 588\\
     relationship   &   37    & judgment  & 720  & video & 44& publication & 579\\
     claim   &   33   & prank  & 719   & said & 33& exclusively & 551\\
    split   &   32    & poised  & 718 &  sexual & 32 & rumor & 537\\
    radar   &   32    & resilient  & 714   & notification & 25& recalls & 535\\
    magazine   &   30    & predicted  & 714   & character & 25 & kardashian & 525\\
    \hline
    \end{tabular}
    \caption{Top ten Literals captured by clauses of TM for GossipCop.}
 \label{explainability_table_gossipcop}
 \vspace{-0.4cm}
\end{table*}
\subsection{Case Study: Interpretability}
We now investigate the interpretability of our TM approach in fake news classification by analyzing the words captured by the clauses for both true and fake news. In brief, the clauses capture two different types of literals:
\begin{itemize}
	\item Plain literals, which are plain words from the target class. 
	\item Negated literals, which are negated words from the other classes.
\end{itemize}
The TM utilizes both plain and negated word patterns for classification. When we analyze the clauses, we see that most of the sub-patterns captured consist of negated literals. This helps TM make decisions robustly because it can use both  negated features from the negative polarity clauses of other classes, as well as the the plain features from positive polarity clauses for the intended class. Taking the positive- and negative polarity claues together, one gets stronger discrimination power.

Table \ref{explainability_table_politifact} and Table \ref{explainability_table_gossipcop} exemplify the above behavior for fake news detection. To showcase the explainability of our approach, we list the ten most captured plain and negated words per class, for both datasets. We observe that negated literals appear quite frequently in the clauses.  This allows the trained TM to represent the class by the features that characterize the class  as well as the features that contrast it from the other class. TM clauses that contain negated and unnegated inputs are termed non-monotone clauses, and these are crucial for human commonsense reasoning, as explored in ``Non-monotonic reasoning'' \cite{38}. 

Also observe that the TM clauses include both descriptive words and discriminative words. For example, in PolitiFact, the \textit{Fake} class captures words like ``trump'', while the \textit{True} class captures the negated version, i.e., ``$\lnot$trump''. However, this does not mean that all the news related to Trump are Fake. Actually, it means that if we break down the clauses into literals, we see “trump” as a descriptor/discriminator for the Fake News class. Therefore, most of the clauses captured the word ``trump'' in plain and in negated form (for the True news class). However, one single word cannot typically produce an accurate classification decision. It is the joint contribution of the literals in a clause that contributes to the high accuracy. Hence, we have to look at all word pattern captured by the clauses. When an input is passed into the trained TM, the clauses from both classes that capture the input word pattern are activated, to vote for their respective class. Finally, the classification is made based on the total votes gathered by both classes for the particular input.

\section{Conclusions}
In this paper, we propose an explainable and interpretable Tsetlin Machine (TM) framework for fake news classification. Our TM framework employs clauses to capture the lexical and semantic features based on word patterns in a document. We also explain the transparent TM learning of clauses from labelled text. The extensive experimental evaluations demonstrate the effectiveness of our model on real-world datasets over various baselines. Our results show that our approach is competitive with far more complex and non-transparent methods, including BERT and XLNet. In addition, we demonstrate how fake news can be ranked according to a credibility score based on classification confidence. We finally demonstrate the explainability of our model using a case study. In our future work, we intend to go beyond using pure text features, also incorporating spatio-temporal and other meta-data features available from social media content, for potentially improved accuracy.

%
%
%
\bibliographystyle{acl_natbib.bst}
\bibliography{ acl2021.bib}

\end{document}